# Model based Bayesian Exploration

**Richard Dearden**
Department of Computer Science
University of British Columbia
Vancouver, BC V6T 1Z4, CANADA
*dearden@cs.ubc.ca*

**Nir Friedman**
Institute of Computer Science
Hebrew University
Jerusalem 91904, ISRAEL
*nir@cs.huji.ac.il*

**David Andre**
Computer Science Division
387 Soda Hall
University of California
Berkeley, CA 94720-1776, USA
*dandre@cs.berkeley.edu*

## Abstract

Reinforcement learning systems are often concerned with balancing exploration of untested actions against exploitation of actions that are known to be good. The benefit of exploration can be estimated using the classical notion of *Value of Information* — the expected improvement in future decision quality arising from the information acquired by exploration. Estimating this quantity requires an assessment of the agent's uncertainty about its current value estimates for states.

In this paper we investigate ways to represent and reason about this uncertainty in algorithms where the system attempts to learn a model of its environment. We explicitly represent uncertainty about the parameters of the model and build probability distributions over Q-values based on these. These distributions are used to compute a myopic approximation to the value of information for each action and hence to select the action that best balances exploration and exploitation.

## 1 Introduction

*Reinforcement learning* addresses the problem of how an agent should learn to act in dynamic environments. This is an important learning paradigm for domains where the agent must consider sequences of actions to be made throughout its lifetime. The framework underlying much of reinforcement learning is that of *Markov Decision Processes* (MDPs). These processes describe the effects of actions in a stochastic environment, and the possible rewards at various states of the environments. If we have an MDP we can compute the choice of actions that maximizes the expected future reward. The task in reinforcement learning is to achieve this level of performance when the underlying MDP is *not* known in advance.

A central debate in reinforcement learning is over the use of models. *Model-free* approaches attempt to learn near-optimal policies without explicitly estimating the dynamics of the surrounding environment. This is usually done by directly approximating a *value function* that measures the desirability of each environment state. On the other hand, *model-based* approaches attempt to estimate a model of the environment's dynamics and use it to compute an estimate of the expected value of actions in the environment.

A common argument for model-based approaches is that by learning a model the agent can avoid costly repetition of steps in the environment. Instead, the agent can use the model to learn the effects of its actions at various states. This can lead to a significant reduction in the number of steps actually executed by the learner, since it can "learn" from simulated steps in the model (Sutton 1990).

Virtually all of the existing model-based approaches in the literature use simple estimation methods to learn the environment, and keep a *point-estimate* of the environment dynamics. Such estimates ignore the agent's uncertainty about various aspects of the environment's dynamics.

In this paper, we advocate a Bayesian approach to model-based reinforcement learning. We show that under fairly reasonable assumptions we can represent the posterior distribution over possible models given our past experience. This is done with essentially the same cost as maintaining point estimates. Our methods thus allow us to continually update this distribution over possible models as we perform actions in the environment.

By representing a distribution over possible models, we can quantify our uncertainty as to what are the best actions to perform. This gives us a handle on the *exploitation vs. exploration* problem. Roughly speaking, this problem involves the dilemma of whether to explore — perform new actions that can lead us to uncharted territories — or to *exploit* — perform actions that have the best performance according to our current knowledge. Clearly, the uncertainty about our model and our expectations as to the range of possible results of actions play crucial roles in this problem.

In a precursor to this work, Dearden et al. (1998) introduce a Bayesian model-free approach in which uncertainty about the Q-values of actions is represented using probability distributions. By explicitly reasoning using uncertainty about Q-values, they direct exploration specifically toward poorly known regions of the state space. Their approach is based on a decision-theoretic approach to action selection: the agent should choose actions based on the value of the information it can expect to learn by performing them (Howard 1966). Dearden et al. propose a measure that balances the expected gains in performance from exploration — in the form of improved policies — with the expected cost of doing a potentially suboptimal action. This measure is computed from probability distributions over the Q-values of actions.

In this paper, we show how to use the posterior distribution over possible models to estimate the distribution of possible Q-values, and then use these to select actions. This use of models allows us to avoid the problem faced by model-free exploration methods, such as the one used by Dearden et al., that need to perform repeated actions to propagate values from one state to another. The main question is how to estimate these Q-values from our distribution of possible models. We present several methods of stochastic sampling to approximate these Q-value distributions. We then evaluate the performance of the resulting Bayesian learning agents on test environments that are designed to fool many exploration methods.

In Section 2 we briefly review the definition of MDPs and the definition of reinforcement learning problems. In Section 3 we discuss a Bayesian approach for learning models. In Section 4 we review the notion of Q-value distributions and the use of value of information for directing exploration and the notion. In Section 5 we propose several sampling methods for estimating Q-value distributions based on the uncertainty about the underlying model. In Section 6 we discuss several approaches of generalizing from the samples we get from the aforementioned methods, and how this generalization can improve our algorithms. In Section 7 we compare our methods to Prioritized Sweeping (Moore & Atkeson 1993), a well known model-based reinforcement learning procedure.

## 2 Background

We assume the reader is familiar with the basic concepts of MDPs (see, e.g., (Kaelbling, Littman & Moore 1996)). We will use the following notation: An MDP is a 4-tuple, $(\mathcal{S}, \mathcal{A}, p_T, p_R)$ where $\mathcal{S}$ is a set of *states*, $\mathcal{A}$ is a set of *actions*, $p_T(s \xrightarrow{a} t)$ is a *transition model* that captures the probability of reaching state $t$ after we execute action $a$ at state $s$, and $p_R(s \xrightarrow{a} r)$ is a *reward model* that captures the probability of receiving reward $r$ after executing $a$ at state $s$. For the reminder of this paper, we assume that possible rewards are a finite subset $\mathcal{R}$ of the real numbers.

In this paper, we focus on infinite-horizon MDPs with a discount factor $\gamma$. The agent's aim is to maximize the expected discounted total reward it receives. Equivalently, we can compute a optimal value function $V^*$ and a Q-function $Q^*$. These functions satisfy the *Bellman equations:*

$$V^*(s) = \max_{a \in \mathcal{A}} Q^*(s, a),$$

where

$$Q^*(s, a) = E_{p_R(s \xrightarrow{a} r)}[r|s, a] + \gamma \sum_{s' \in \mathcal{S}} p_T(s \xrightarrow{a} s') V^*(s').$$

If the agent has access to $V^*$ or $Q^*$, it can optimize its expected reward by choosing the action $a$ at $s$ that maximizes $Q^*(s, a)$. Given a model, we can compute $Q^*$ using a variety of methods, including *value iteration*. In this method we repeatedly update an estimate $Q$ of $Q^*$ by applying the Bellman equations to get new values of $Q(s)$ for some (or all) of the states.

Reinforcement learning procedures attempt to achieve an optimal policy when the agent *does not* know $p_T$ and $p_R$. Since we do not know the dynamics of the underlying MDP, we cannot compute the Q-value function directly. However, we can estimate it. In *model-free* approaches one usually estimates $Q$ by treating each step in the environment as a sample from the underlying dynamics. These samples are then used for performing updates of the Q-values based on the Bellman equations. In *model-based* reinforcement learning one usually directly estimates $p_T(s \xrightarrow{a} t)$ and $p_R(s \xrightarrow{a} r)$. The standard approach is then to act as though these approximations are correct, compute $Q^*$, and use it to choose actions.

A standard problem in learning is balancing between planning (i.e., choosing a policy) and execution. Ideally, the agent would compute the optimal value function for its model of the environment each time it updates it. This scheme is unrealistic since finding the optimal policy for a given model is a non-trivial computational task. Fortunately, we can approximate this scheme if we notice that the approximate model changes only slightly at each step. We can hope that the value function from the previous model can be easily "repaired" to reflect these changes. This approach was pursued in the DYNA (Sutton 1990) framework, where after the execution of an action, the agent updates its model of the environment, and then performs some bounded number of value propagation steps to update its approximation of the value function. Each value-propagation step locally enforces the Bellman-equation by setting $\hat{V}(s) \leftarrow \max_{a \in \mathcal{A}} \hat{Q}(s, a)$, where $\hat{Q}(s, a) = E[\hat{p}_R(s \xrightarrow{a} r)] + \gamma \sum_{s' \in \mathcal{S}} \hat{p}_T(s \xrightarrow{a} s') \hat{V}(s')$, $\hat{p}_T(s \xrightarrow{a} s')$ and $\hat{p}_R(s \xrightarrow{a} r)$ are the agent's approximate model, and $\hat{V}$ is the agent's approximation of the value function.

This raises the question of which states should be updated. Prioritized Sweeping (Moore & Atkeson 1993) is a method that estimates to what extent states would change their value as a consequence of new knowledge of the MDP dynamics or previous value propagations. States are assigned priorities based on the expected size of changes in their values, and states with the highest priority are the ones for which we perform value propagation.

## 3 Bayesian Model Learning

In this section we describe how to maintain a Bayesian posterior distribution over MDPs given our experiences in the environment. At each step in the environment, we start at state $s$, choose an action $a$, and then observe a new state $t$ and a reward $r$. We summarize our experience by a sequence of *experience tuples* $\langle s, a, r, t \rangle$.

A Bayesian approach to this learning problem is to maintain a *belief state* over the possible MDPs. Thus, a belief state $\mu$ defines a probability density $P(M \mid \mu)$. Given an experience tuple $\langle s, a, r, t \rangle$ we can compute the *posterior* belief state, which we denote $\mu \circ \langle s, a, r, t \rangle$, by Bayes rule:

$$P(M \mid \mu \circ \langle s, a, r, t \rangle)$$
$$\propto P(\langle s, a, r, t \rangle \mid M) P(M \mid \mu)$$
$$= P(s \xrightarrow{a} t \mid M) P(s \xrightarrow{a} r \mid M) P(M \mid \mu).$$

Thus, the Bayesian approach starts with some *prior* probability distribution over all possible MDPs (we assume that the sets of possible states, actions and rewards are delimited in advance). As we gain experience, the approach focuses the mass of the *posterior* distribution on those MDPs in which the observed experience tuples are most probable.

An immediate question is whether we can represent these prior and posterior distributions over an infinite number of MDPs. We show that this is possible by adopting results from Bayesian learning of probabilistic models, such as Bayesian networks (Heckerman 1998). Under carefully chosen assumptions, we can represent such priors and posteriors in any of several compact manners. We discuss one such choice below.

To formally represent our problem, we consider the *parameterization* of MDPs. The simplest parameterization is table based, where there are parameters $\theta^t_{s,a,t}$ and $\theta^r_{s,a,r}$ for the transition and reward models. Thus, for each choice of $s$ and $a$, the parameters $\theta^t_{s,a} = \{\theta^t_{s,a,t} : t \in \mathcal{S}\}$ define a distribution over possible states, and the parameters $\theta^r_{s,a} = \{\theta^r_{s,a,r} : r \in \mathcal{R}\}$ define a distribution over possible rewards.[1]

We say that our prior satisfies *parameter independence* if it has the product form:

$$\Pr(\theta \mid \mu) = \prod_s \prod_a \Pr(\theta^t_{s,a} \mid \mu) \Pr(\theta^r_{s,a} \mid \mu). \quad (1)$$

Thus, the prior distribution over the parameters of each local probability term in the MDP is independent of the prior over the others. It turns out that this form is maintained as we incorporate evidence.

**Proposition 3.1:** *If the belief state $P(\theta \mid \mu)$ satisfies parameter independence, then $P(\theta \mid \mu \circ \langle s, a, r, t \rangle)$ also satisfies parameter independence.*

As a consequence, the posterior after we incorporate an arbitrarily long number of experience tuples also has the product form of (1).

Parameter independence allows us to reformulate the learning problem as a collection of unrelated local learning problems. In each of these, we have to estimate a probability distribution over all states or all rewards. The question is how to learn these distributions. We can use well-known Bayesian methods for learning standard distributions such as multinomials or Gaussian distributions (Degroot 1986).

For the case of discrete multinomials, which we have assumed in our transition and reward models, we can use *Dirichlet* priors to represent $\Pr(\theta^t_{s,a})$ and $\Pr(\theta^r_{s,a})$. These priors are *conjugate*, and thus the posterior after each observed experience tuple will also be a Dirichlet distribution. In addition, Dirichlet distributions can be described using a small number of *hyper-parameters*. See Appendix A for a review of Dirichlet priors and their properties.

In the case of most MDPs studied in reinforcement learning, we expect the transition model to be sparse—there are only a few states that can result from a particular action at a particular state. Unfortunately, if the state space is large, learning with a Dirichlet prior can require many examples to recognize that most possible states are highly unlikely. This problem is addressed by a recent method of learning sparse-multinomial priors (Friedman & Singer 1999). Without going into details, the sparse-multinomial priors have the same general properties as Dirichlet priors, but assume that the observed outcomes are from some small subsets of the set of possible ones. The sparse Dirichlet priors make predictions as though only the observed outcomes are possible, except that they also assign to *novel* outcomes. In the MDP setting, a novel outcome is a transition to state $t$ that was not reached from $s$ previously by executing $a$. See Appendix A for a brief summary of sparse-multinomial priors and their properties.

For both the Dirichlet and its sparse-multinomial extension, we need to maintain the number of times, $N(s \xrightarrow{a} t)$, state $t$ is observed after executing action $a$ at state $s$, and similarly, $N(s \xrightarrow{a} r)$ for rewards. With the prior distributions over the parameters of the MDP, these counts define a posterior distribution over MDPs. This representation allows us to both predict the probability of the next transition and reward, and also to compute the probability of every possible MDP and to sample from the distribution of MDPs.

To summarize, we assumed parameter independence, and that for each prior in (1) we have either a Dirichlet or sparse-multinomial prior. The consequence is that the posterior at each stage in the learning can be represented compactly. This enables us to estimate a distribution over MDPs at each stage.

It is easy to extend this discussion for more compact parameterizations of the transition and reward models. For example, if each state is described by several attributes, we might use a Bayesian network to capture the MDP dynamics. Such a structure requires fewer parameters and thus we can learn it with fewer examples. Nonetheless, much of the above discussion and conclusions about parameter independence and Dirichlet priors apply to these models (Heckerman 1998).

Standard model-based learning methods maintain a point estimate of the model. These point estimates are often close to the *mean* prediction of the Bayesian method. However, these point estimates do not capture the uncertainty about the model. In this paper, we examine how knowledge of this uncertainty can be exploited to improve exploration.

---

[1] The methods we describe are easily extend to other parameterizations. In particular, we can consider continuous distributions, e.g., Gaussians, over rewards. For clarity of discussion, we focus on multinomial distributions throughout the paper.

## 4  Value of Information Exploration

In a recent paper, Dearden et al. (1998) examined model-free Bayesian reinforcement learning. Their approach builds on the notion of *Q-value distributions*. Recall, that $Q^*(s, a)$ is the expected reward if we execute $a$ at $s$ and then continue with optimal selection of actions. Since during learning we are uncertain about the model, there is a distribution over the Q-values at each pair $(s, a)$. This distribution is induced by the belief state over possible MDPs, and the Q-values for each of these MDPs. In the model-free case, Dearden et al. propose an approach for estimating Q-value distributions without building a model. This approach makes several strong assumptions that are clearly violated in MDPs. In the next section, we show how we can use our representation of the posterior over models to give estimates of Q-value distributions. Before we do that, we briefly review how Dearden et al. use the Q-value distributions for selecting actions, as we use this method in the current work.

The approach of Dearden et al is based on the decision-theoretic ideas of *value of information* (Howard 1966). The application of these ideas in this context is reminiscent of its use in tree search (Russell & Wefald 1991), which can also be seen as a form of exploration. The idea is to balance the expected gains from exploration—in the form of improved policies—against the expected cost of doing a potentially suboptimal action.

To formally define the approach, we need to introduce some notation. We denote by $q_{s,a}$ a possible value of $Q^*(s, a)$ in some MDP. We treat these quantities as random variables that depend on our belief state. (For clarification of the following discussion, we do not explicitly reference the belief state in the mathematical notation.) We now consider what can be gained by learning the true value $q^*_{s,a}$ of $q_{s,a}$. How would this knowledge change the agent's future rewards? Clearly, if this knowledge does not change the agent's policy, then future rewards would not change. Thus, the only interesting scenarios are those where the new knowledge does change the agent's policy. This can happen in two cases: (a) when the new knowledge shows that an action previously considered sub-optimal is revealed as the best choice (given the agent's beliefs about other actions), and (b) when the new knowledge indicates that an action that was previously considered best is actually inferior to other actions.

For case (a), suppose that $a_1$ is the best action; that is, $E[q_{s,a_1}] \geq E[q_{s,a'}]$ for all other actions $a'$. Moreover suppose that the new knowledge indicates that $a$ is a better action; that is, $q^*_{s,a} > E[q_{s,a_1}]$. Thus, we expect the agent to gain $q^*_{s,a} - E[q_{s,a_1}]$ by virtue of performing $a$ instead of $a^*$.

For case (b), suppose that $a_1$ is the action with the highest expected value and $a_2$ is the second-best action. If the new knowledge indicates that $q_{s,a_1} < E[q_{s,a_2}]$, then the agent should perform $a_2$ instead of $a_1$ and we expect it to gain $E[q_{s,a_2}] - q^*_{s,a_1}$.

Combining these arguments, the *gain* from learning the value of $q^*_{s,a}$ of $q_{s,a}$ is:

$$Gain_{s,a}(q^*_{s,a}) = \begin{cases} E[q_{s,a_2}] - q^*_{s,a} & \text{if } a = a_1 \text{ and } q^*_{s,a} < E[q_{s,a_2}] \\ q^*_{s,a} - E[q_{s,a_1}] & \text{if } a \neq a_1 \text{ and } q^*_{s,a} > E[q_{s,a_1}] \\ 0 & \text{otherwise} \end{cases}$$

where, again, $a_1$ and $a_2$ are the actions with the best and second best expected values respectively. Since the agent does not know in advance what value will be revealed for $q^*_{s,a}$, we need to compute the *expected* gain given our prior beliefs. Hence the expected value of perfect information about $q_{s,a}$ is:

$$VPI(s, a) = \int_{-\infty}^{\infty} Gain_{s,a}(x) \Pr(q_{s,a} = x) dx \quad (2)$$

The computation of this integral depends on how we represent our distributions over $q_{s,a}$. We return to this issue below.

The value of perfect information gives an upper bound on the myopic value of information for exploring action $a$. The expected *cost* incurred for this exploration is given by the difference between the value of $a$ and the value of the current best action, i.e., $\max_{a'} E[q_{s,a'}] - E[q_{s,a}]$. This suggests we choose the action that maximizes

$$VPI(s, a) - (\max_{a'} E[q_{s,a'}] - E[q_{s,a}]).$$

Clearly, this strategy is equivalent to choosing the action that maximizes:

$$E[q_{s,a}] + VPI(s, a).$$

We see that the value of exploration estimate is used as a way of boosting the desirability of different actions. When the agent is confident of the estimated $Q$-values, the VPI of each action is close to 0, and the agent will always choose the action with the highest expected value.

## 5  Estimating Q-Value Distributions

How do we estimate the Q-value distributions? We now examine several methods of different complexity and bias.

### 5.1  Naive Sampling

Perhaps the simplest approach is to simulate the definition of a Q-value distribution. Since there are an infinite number of possible MDPs, we cannot afford to compute Q-values for each. Instead, we sample $k$ MDPs: $M^1, \ldots, M^k$ from the distribution $\Pr(M \mid \mu)$. We can solve each MDP using standard techniques (e.g., value iteration or linear programming). For each state $s$ and action $a$, we then have a sample solution $q^1_{s,a}, \ldots, q^k_{s,a}$, where $q^i_{s,a}$ is the optimal Q-value, $Q^*(s, a)$, given the $i$'th MDP. From this sample we can estimate properties of the Q-distribution. For generality, we denote the weight of each sample, given $\mu$, as $w^i_\mu$. Initially these weights are all equal to 1.

Given these samples, we can estimate the mean Q-value as

$$E[q_{s,a}] \approx \frac{1}{\sum_i w_\mu^i} \sum_i w_\mu^i q_{s,a}^i.$$

Similarly, we can estimate the VPI by summing over the $k$ MDPs:

$$VPI(s, a) \approx \frac{1}{\sum_i w_\mu^i} \sum_i w_\mu^i Gain_{s,a}(q_{s,a}^i).$$

This approach is straightforward; however, it requires an efficient sampling procedure. Here again the assumptions we made about the priors helps us. If our prior has the form of (1), then we can sample each distribution ($p_T(s \xrightarrow{a} t)$ or $p_R(s \xrightarrow{a} r)$) independently of the rest. Thus, the sampling problem reduces to sampling from "simple" posterior distributions. For Dirichlet priors there are known sampling methods. For the sparse-multinomials the problem is a bit more complex, but solvable. In Appendix A we describe both sampling methods.

### 5.2 Importance Sampling

An immediate problem with the naive sampling approach is that it requires several global computations (e.g., computing value functions for MDPs) to evaluate each action made by the agent. This is clearly too expensive. One possible way of avoiding these repeated computations is to reuse the same sampled MDPs for several steps. To do so, we can use ideas from *importance sampling*.

In importance sampling we want to a sample from $\Pr(M \mid \mu')$ but for some reasons, we actually sample from $\Pr(M \mid \mu)$. We adjust the weight of each sample appropriately to correct for the difference between the *sampling* distribution (e.g., $\Pr(M \mid \mu)$) and the *target* distribution (e.g., $\Pr(M \mid \mu')$):

$$w_{\mu'}^i = \frac{\Pr(M^i \mid \mu')}{\Pr(M^i \mid \mu)} w_\mu^i.$$

We now use the weighted sum of samples to estimate the mean and the VPI of different actions. It is easy to verify that the weighted sample leads to correct prediction when we have a large number of samples. In practice, the success of importance sampling depends on the difference between the two distributions. If an MDP $M$ has low probability according to $\Pr(M \mid \mu)$, then the probability of sampling it is small, even if $\Pr(M \mid \mu')$ is high.

Fortunately for us, the differences between the beliefs before and after observing an experience tuple are usually small. We can easily show that

**Proposition 5.1:**

$$\begin{aligned} w_{\mu \circ (s,a,r,t)} &= \frac{\Pr(M \mid \mu \circ \langle s,a,r,t \rangle)}{\Pr(M \mid \mu)} w_\mu \\ &= \frac{\Pr(\langle s,a,r,t \rangle \mid M)}{\Pr(\langle s,a,r,t \rangle \mid \mu)} w_\mu \end{aligned}$$

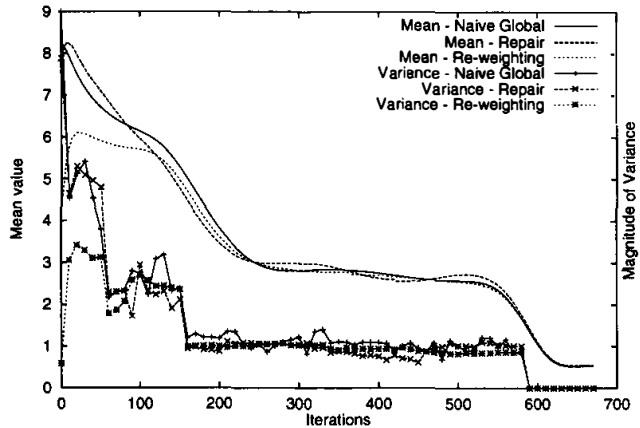

Figure 1: Mean and variance of the Q-value distribution for a state, plotted as a function of time. Note that the means of each method converge to the true value of the state at the same time that the variances approach zero.

The term $\Pr(\langle s, a, r, t \rangle \mid M)$ is easily computed from $M$, and $\Pr(\langle s, a, r, t \rangle \mid \mu)$ can be easily computed based on our posteriors. Thus, we can easily re-weight the sampled models after each experience is recorded and use the weighted sum for choosing actions. Note that re-weighting of models is fast, and since we have already computed the Q-value for each pair $(s, a)$ in each of the models, no additional computations are needed.

Of course, the original set of models we sampled becomes irrelevant as we learn more about the underlying MDP. We can use the total weight of the sampled MDPs to track how unlikely they are given the observations. Initially this weight is $k$. As we learn more it usually becomes smaller. When it becomes smaller than some threshold $k_{\min}$, we sample $k - k_{\min}$ new MDPs from our current belief state, assigning each one weight 1 and thus bringing the total weight of the sample to $k$ again. We then need only to solve the newly sampled MDPs.

To summarize, we sample $k$ MDPs, solve them, and use the $k$ Q-values to estimate properties of the Q-value distribution. We re-weight the samples at each step to reflect our newly gained knowledge. Finally, we have an automatic method for detecting when new samples are required.

### 5.3 Global Sampling with Repair

The global sampling approach of the previous section has one serious deficiency. It involves computing global solutions to MDPs which can be very expensive. Although we can reuse MDPs from previous steps, this approach still requires us to sample new MDPs and solve them quite often.

An alternative idea is to keep updating each of the sampled MDPs. Recall that after observing an experience tuple $\langle s, a, r, t \rangle$, we only change the posterior over $\theta_{s,a}^t$ and $\theta_{s,a}^r$. Thus, instead of re-weighting the sample $M^i$, we can update, or *repair*, it by re-sampling $\theta_{s,a}^t$ and $\theta_{s,a}^r$. If the orig-inal sample $M^i$ was sampled from $\Pr(M \mid \mu)$, then it easily follows that the repaired $M^i$ is sampled from $\Pr(M \mid \mu \circ \langle s, a, r, t \rangle)$.

Of course, once we modify $M^i$ its Q-value function changes. However, all of these changes are consequences of the new values of the dynamics at $(s, a)$. Thus, we can use prioritized sweeping to update the Q-value computed for $M^i$. This sweeping performs several Bellman updates to correct the values of states that are affected by the change in the model.[2]

This suggests the following algorithm. Initially, we sample $k$ MDPs from our prior belief state. At each step we:

- Observe an experience tuple $\langle s, a, r, t \rangle$
- Update $\Pr(\theta^t_{s,a})$ by $t$, and $\Pr(\theta^r_{s,a})$ by $r$.
- For each $i = 1, \ldots, k$, sample $\theta^{t,i}_{s,a}$, $\theta^{r,i}_{s,a}$ from the new $\Pr(\theta^t_{s,a})$ and $\Pr(\theta^r_{s,a})$, respectively.
- For each $i = 1, \ldots, k$ run a local instantiation of prioritized sweeping to update the Q-value function of $M^i$.

Thus, our approach is quite similar to standard model based learning with prioritized sweeping, but instead of running one instantiation of prioritized sweeping, we run $k$ instantiations in parallel, one for each sampled MDP. The repair to the sampled MDPs ensures that they constitute a sample from the current belief state, and the local instantiations of prioritized sweeping ensure that the Q-values computed in each of these MDPs is a good approximation to the true value.

As with the other approaches we have described, after we invoke the $k$ prioritized sweeping instances we use the $k$ samples from each $q_{s,a}$ to select the next actions using VPI computations.

Figure 1 shows a single run of learning where the actions selected were fixed and each of the three methods was used to estimate the Q-values of a state. Initially the means and variances are very high, but as the agent gains more experience, the means converge on the true value of the state, and the variances tend towards zero. These results suggest that the repair and importance sampling approaches both provide reasonable approximations to naive global sampling.

### 5.4 Local Sampling

Until now we have considered using global samples of MDPs. An alternative approach is to try to maintain for each $(s, a)$ an estimate of the Q-value distribution, and to update these distributions using a local, Bellman-update like, propagation rule. To understand this approach, recall the Bellman equation:

$$q_{s,a} = E[p_R(s \xrightarrow{a} r)] + \gamma \sum_{s' \in \mathcal{S}} p_T(s \xrightarrow{a} s') \max_{a'} q_{s',a'}.$$

---
[2]Generalized prioritized sweeping (Andre, Friedman & Parr 1997) allows us to extend prioritized sweeping to these approximate settings. When using approximate models or value functions, one must address the problem of calculating the states on which to estimate the priority.

In our current setting, the terms $q_{s',a'}$ are random variables that depend on our current estimate of Q-value distributions. The probabilities $p_T(s \xrightarrow{a} s')$ are also random variables that depend on our posterior on $\theta^t_{s,a}$, and finally $E[p_R(s \xrightarrow{a} r)]$ is also a random variable that depends on the posterior on $\theta^r_{s,a}$. Thus, we can sample from $q_{s,a}$, by jointly sampling from all of these distributions, i.e., $q_{s',a'}$ for all states, $p_T(s \xrightarrow{a} s')$, and $p_R(s \xrightarrow{a} r)$., and then computing the Q-value. If we repeat this sampling step $k$ times, we get $k$ samples from a single bellman iteration for $q_{s,a}$.

Starting with our beliefs about the model and about the Q-value distribution of all states, we can sample from the distribution of $q_{s,a}$. To make this procedure manageable, we assume that we can sample from each $q_{s',a'}$ independently. This assumption does not hold in general MDPs, since the distribution of different Q-values are correlated (by the Bellman equation). However, we might hope that the exponential decay will weaken these dependencies.

We are now left with the question how to use the $k$ samples from $q_{s,a}$. The simplest approach is to use the samples as a representation of our approximation of the distribution of $q_{s,a}$. We can compute the mean and VPI from a set of samples, as we did in the global sampling approach. Similarly, we can re-sample from this representation by randomly choosing one of the points. This results in a method that is similar to recent sampling methods that have been used successfully in monitoring complex dynamic processes (Kanazawa, Koller & Russell 1995).

This gives us a method for performing a Bellman-update on our Q-value distributions. To get a good estimate of these distributions we need to repeat these updates. Here we can use a prioritized sweeping like algorithm that performs updates based on an estimate of which Q-value distribution can be most affected by the updates to other Q-value distributions.

## 6 Generalization and Smoothing

In the approaches described above we generated samples from the Q-value distributions, and effectively used a collection of points to represent the approximation to the Q-Value distribution. A possible problem with this representation approach is that we use a fairly simplistic representation to describe a complex distribution. This suggests that we should generalize from the $k$ samples by using standard generalization methods.

This is particularly important in the local sampling approach. Here we also use our representation of the Q-value distribution to propagate samples for other Q-value distributions. Experience from monitoring tasks in stochastic processes suggest that introducing generalization can drastically improve performance (Koller & Fratkina 1998).

Perhaps the simplest approach to generalize from the $k$ samples is to assume that the Q-value distribution has a particular parametric form, and then to fit the parameters to the samples. The first approach that comes to mind is fitting a Gaussian to the $k$ samples. This captures the first two moments of the sample, and allows simple generalization. Unfortunately, because of the max() terms in the Bellman equations, we expect the Q-value distribution to be skewed to the positive direction. If this skew is strong, then fitting a Gaussian would be a poor generalization from the sample.

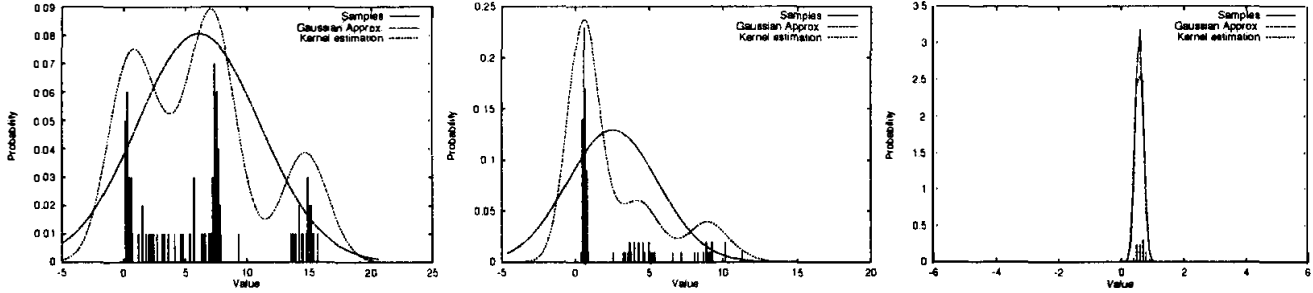

Figure 2: Samples, Gaussian approximation, and Kernel estimates of a Q-value distribution after 100, 300, and 700 steps of Naive global sampling on the same run as Figure 1.

At the other end of the spectrum are non-parametric approaches. One of the simplest ones is *Kernel estimation* (see for example (Bishop 1995)). In this approach, we approximate the distribution over $Q(s, a)$ by a sum of Gaussians with a fixed variance, one for each sample. This approach can be effective if we are careful in choosing the variance parameter. A too small variance, will lead to a spiky distribution, a too large variance, will lead to an overly smooth and flat distribution. We use a simple rule for estimating the kernel width as a function of the mean (squared) distance between points.[3]

Of course, there are many other generalization methods we might consider using here, such as mixture distributions. However, these two approaches provide us with initial ideas on the effect of generalization in this context.

We must also compute the VPI of a set of generalized distributions made up of Gaussians or kernel estimates. This is simply a matter of solving the integral given in Equation 2

---

[3]This rule is motivated by a *leave-one-out cross-validation* estimate of the kernel widths. Let $q^1, \ldots, q^k$ be the $k$ samples. We want to find the kernel width $\sigma$ that maximizes the term

$$J(\sigma^2) = \sum_i \log(\sum_{j \neq i} f(q^i | q^j, \sigma^2))$$

where $f(q^i | q^j, \sigma)$ is the Gaussian pdf with mean $q^j$ and variance $\sigma^2$. Using Jensen's inequality, we have that

$$J(\sigma^2) \geq \sum_i \sum_{j \neq i} \log f(q^i | q^j, \sigma^2)$$

**Proposition 6.1** : *The value of $\sigma^2$ that maximizes $\sum_i \sum_{j \neq i} \log f(q^i | q^j, \sigma^2)$ is $\frac{1}{4}d$, where $d$ is the average distance among samples:*

$$d = \frac{1}{k(k-1)} \sum_i \sum_{j \neq i} (q^i - q^j)^2$$

where $\Pr(q_{s,a} = x)$ is computed from the generalized probability distribution for state $s$ and action $a$. This integration can be simplified to a term where the main cost is an evaluation of the cdf of a Gaussian distribution (e.g., see (Russell & Wefald 1991). This function, however, is implemented in most language libraries (e.g., using the *erf*() function in the C-library), and thus can be done quite efficiently.

Figure 2 shows the effects of Gaussian approximation and kernel estimation smoothing (using the computed kernel width) on the sample values used to generate the Q-distributions in Figure 1 for three different time steps. Early in the run Gaussian approximation produces a very poor approximation because the samples are quite widely spread and very skewed, while kernel estimation provides a much better approximation to the observed distribution. For this reason, we expect kernel estimation to perform better than Gaussian approximation for computing VPI.

## 7 Experimental Results

Figure 3 shows two domains of the type on which we have tested our algorithms. Each is a four action maze domain in which the agent begins at the point marked $S$ and must collect the flag $F$ and deliver it to the goal $G$. The agent receives a reward of 1 for each flag it collects and then moves to the goal state, and the problem is then reset. If the agent enters the square marked $T$ (a trap) it receives a reward of -10. Each action (up, down, left, right) succeeds with probability 0.9 if that direction is clear, and with probability 0.1, moves the agent perpendicular to the desired direction. The "trap" domain has 18 states, the "maze" domain 56.

We evaluate the algorithms by computing the average (over 10 runs) future discounted reward received by the agent. We use this measure rather than the value of the learned policy because exploratory agents rarely actually follow either the greedy policy they have discovered or their current exploration policy for very long. For comparison we use prioritized sweeping (Moore & Atkeson 1993) with the $T_{\text{bored}}$ parameter optimized for each problem.

Figure 4 shows the performance of a representative sample of our algorithms on the trap domain. Unless they are based on a very small number of samples, all of the Bayesian exploration methods outperform prioritized sweeping. This is due to their more cautious approach to the trap state. Although they are uncertain about it, they know that its value is probably bad, and hence do not explore it further after a small number of visits.

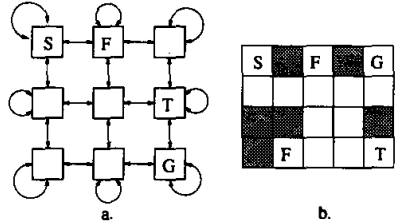

Figure 3: The (a.) "trap" and (b.) larger maze domains.

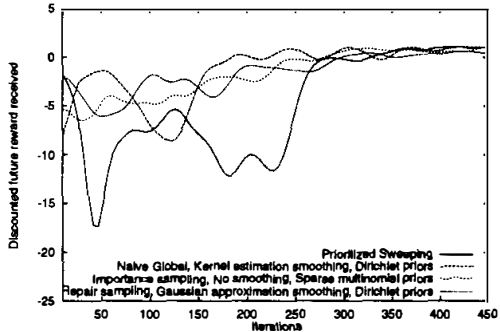

Figure 4: Discounted future reward received for the "trap" domain.

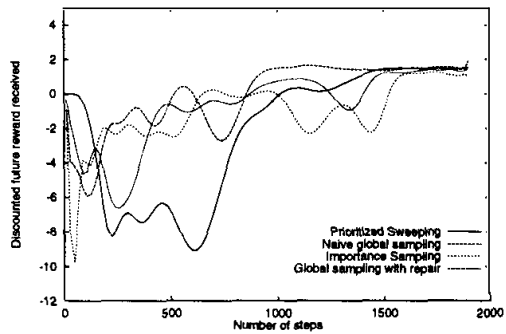

Figure 5: Comparison of Q-value estimation techniques on the larger maze domain.

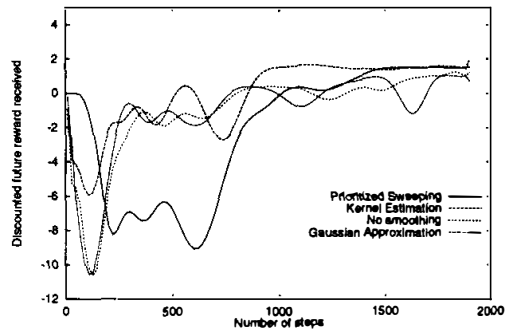

Figure 6: The effects of smoothing techniques on performance in the large maze domain.

Figure 5 compares prioritized sweeping with our Q-value estimation techniques on the larger maze domain. As the graph shows, our techniques perform better than prioritize sweeping early in the learning process. They explore more widely initially, and do a better job of avoiding the trap state once they find it. Of the three techniques, global sampling performs best, although its computational requirements are considerable — about ten times as much as sampling with repair. Importance sampling runs about twice as fast as global sampling but converges relatively late on this problem, and did not converge on all trials.

Figure 6 shows the relative performance of the three smoothing methods, again on the larger domain. To exaggerate the effects of smoothing, only 20 samples were used to produce this graph. Kernel estimation performs very well, while no smoothing failed to find the optimal (two flag) strategy on two out of ten runs. Gaussian approximation was slow to settle on a policy, it continued to make exploratory actions after 1500 steps while all the other algorithms had converged by then.

We are currently investigating the performance of the algorithm on both more complex maze domains and random MDPS, and also the effectiveness of the local sampling approach we have described.

## 8 Discussion

This paper makes two main contributions. First, we show how to maintain Bayesian belief states about MDPs. We show that this can be done in a simple manner by using ideas that appear in Bayesian learning of probabilistic models. Second, we discuss how to use the Bayesian belief state to choose actions in a way that balances exploration and exploitation. We adapt the value of information approach of Dearden et al. (1998) to this model-based setup and show how to approximate the Q-value distributions needed for making these choices.

A recent approach to exploration that is related to our work is that of Kearns and Singh (1998). Their approach divides the set of states in to two groups. The *known* states are ones for which the learner is quite confident about the transition probabilities. That is, the learner believes that its estimate of the transition probabilities is close enough to the true distribution. All other states are considered *unknown* states. In Kearns and Singh's proposal, the learner constructs a policy over the known states. This policy takes into account both exploitation and the possibility of finding better rewards in unknown states (which are considered as highly-rewarding). When it finds itself in an unknown state, the agent chooses actions randomly. The algorithm proceeds in phases, after each one it reclassifies the states and recomputes the policy on the known states. Kearns and Singh's proposal is significant in that it is the first one for which we have polynomial guarantees on number of steps needed to get to a good policy. However, this algorithm was not implemented or tested, and it is not clear how fast it learns in real domains.

Our exploration strategy also keeps a record of how confident we are in each state (i.e., Bayesian posterior), and also chooses actions based on their expected rewards (both known rewards, and possible exploration rewards). The main difference is that we do not commit to a binary classification of states, but instead choose actions in a way that takes into account the possible value of doing the exploration. This leads to exploitation, even before we are extremely confident in the dynamics at every state in the "interesting" parts of the domain.

There are several directions for future research. First, we are currently conducting experiments on larger domains to show how our method scales up. We are also interested in applying it to more compact model representations (e.g., using *dynamic Bayesian networks*), and to problems with continuous state spaces.

Finally, the most challenging future direction is to deal with the actual value of information of an action rather than myopic estimates. This problem can stated as an MDP over belief states. However, this MDP is extremely large, and requires some approximations to find good policies quickly. Some of the ideas we introduced here, such as the re-weighting of sampled MDPs might allow us to address this computational task.

### Acknowledgements

We are grateful for useful comments from Craig Boutilier and Stuart Russell. Richard Dearden was supported by a Killam Predoctoral fellowship and by IRIS Phase-III project "Dealing with Actions" (BAC). Some of this work was done while Nir Friedman was at U.C. Berkeley. Nir Friedman and David Andre were supported in part by ARO under the MURI program "Integrated Approach to Intelligent Systems", grant number DAAH04-96-1-0341, and by ONR under grant number N00014-97-1-0941. Nir Friedman was also supported through the generosity of the Michael Sacher Trust. David Andre was also supported by a DOD National Defense Science and Engineering Grant.

## A    Dirichlet and Sparse-Multinomial Priors

Let $X$ be a random variable that can take $L$ possible values from a set $\Sigma$. Without loss of generality, let $\Sigma = \{1, \ldots L\}$. We are given a training set $D$ that contains the outcomes of $N$ independent draws $x^1, \ldots, x^N$ of $X$ from an unknown multinomial distribution $P^*$. The *multinomial estimation* problem is to find a good approximation for $P^*$.

This problem can be stated as the problem of predicting the outcome $x^{N+1}$ given $x^1, \ldots, x^N$. Given a prior distribution over the possible multinomial distributions, the Bayesian estimate is:

$$P(x^{N+1} \mid x^1, \ldots, x^N, \xi)$$

$$= \int P(x^{N+1} \mid \theta, \xi) P(\theta \mid x^1, \ldots, x^N, \xi) d\theta \quad (3)$$

where $\theta = \langle \theta_1, \ldots, \theta_L \rangle$ is a vector that describes possible values of the (unknown) probabilities $P^*(1), \ldots, P^*(L)$, and $\xi$ is the "context" variable that denote all other assumptions about the domain.

The posterior probability of $\theta$ can be rewritten as:

$$P(\theta \mid x^1, \ldots, x^N, \xi) \propto P(x^1, \ldots, x^N \mid \theta, \xi) P(\theta \mid \xi)$$
$$= P(\theta \mid \xi) \prod_i \theta_i^{N_i}, \quad (4)$$

where $N_i$ is the number of occurrences of the symbol $i$ in the training data.

*Dirichlet* distributions are a parametric family that is *conjugate* to the multinomial distribution. That is, if the prior distribution is from this family, so is the posterior. A Dirichlet prior for $X$ is specified by *hyper-parameters* $\alpha_1, \ldots, \alpha_L$, and has the form:

$$P(\theta \mid \xi) \propto \prod_i \theta_i^{\alpha_i - 1} \quad (\sum_i \theta_i = 1 \text{ and } \theta_i \geq 0 \text{ for all } i)$$

where the proportion depends on a normalizing constant that ensures that this is a legal density function (i.e., integral of $P(\theta \mid \xi)$ over all parameter values is 1). Given a Dirichlet prior, the initial prediction for each value of $X$ is

$$P(X^1 = i \mid \xi) = \int \theta_i P(\theta \mid \xi) d\theta = \frac{\alpha_i}{\sum_j \alpha_j}$$

It is easy to see that, if the prior is a Dirichlet prior with hyper-parameters $\alpha_1, \ldots, \alpha_L$, then the posterior is a Dirichlet with hyper-parameters $\alpha_1 + N_1, \ldots, \alpha_L + N_L$. Thus, we get that the prediction for $X^{N+1}$ is

$$P(X^{N+1} = i \mid x^1, \ldots, x^N, \xi) = \frac{\alpha_i + N_i}{\sum_j (\alpha_j + N_j)}.$$

In some situations we would like to sample a vector $\theta$ according to the distribution $P(\theta \mid \xi)$. This can be done using a simple procedure: Sample values $y_1, \ldots, y_L$ such that each $y_i \sim Gamma(\alpha_i, 1)$ and then normalize to get a probability distribution, where $Gamma(\alpha, \beta)$ is the Gamma distribution. Procedures for sampling from these distributions can be found in (Ripley 1987).

Friedman and Singer (1999) introduce a structured prior that captures our uncertainty about the set of "feasible" values of $X$. Define a random variable $V$ that takes values from the set $2^\Sigma$ of possible subsets of $\Sigma$. The intended semantics for this variable, is that if we know the value of $V$, then $\theta_i > 0$ iff $i \in V$.

Clearly, the hypothesis $V = \Sigma'$ (for $\Sigma' \subseteq \Sigma$) is consistent with training data only if $\Sigma'$ contains all the indices $i$ for which $N_i > 0$. We denote by $\Sigma^o$ the set of observed symbols. That is, $\Sigma^o = \{i : N_i > 0\}$, and we let $k^o = |\Sigma^o|$. Suppose we know the value of $V$. Given this assumption, we can define a Dirichlet prior over possible multinomial

Model based Bayesian Exploration    159distributions $\theta$ if we use the same hyper-parameter $\alpha$ for each symbol in $V$. Formally, we define the prior:

$$P(\theta|V) \propto \prod_{i \in V} \theta_i^{\alpha-1} \quad (\sum_i \theta_i = 1 \text{ and } \theta_i = 0 \text{ for all } i \notin V) \quad (5)$$

Using Eq. (4), we have that:

$$P(X^{N+1} = i \mid x^1, \ldots, x^n, V) = \begin{cases} \frac{\alpha + N_i}{|V|\alpha + N} & \text{if } i \in V \\ 0 & \text{otherwise} \end{cases} \quad (6)$$

Now consider the case where we are uncertain about the actual set of feasible outcomes. We construct a two tiered prior over the values of $V$. We start with a prior over the size of $V$, and assume that all sets of the same cardinality have the same prior probability. We let the random variable $S$ denote the cardinality of $V$. We assume that we are given a distribution $P(S = k)$ for $k = 1, \ldots, L$. We define the prior over sets to be $P(V \mid S = k) = \binom{L}{k}^{-1}$. This prior is a sparse-multinomial with parameters $\alpha$ and $\Pr(S = k)$.

Friedman and Singer show that how we can efficiently predict using this prior.

**Theorem A.1:** (Friedman & Singer 1999) *Given a sparse-multinomial prior, the probability of the next symbol is*

$$P(X^{N+1} = i \mid D) = \begin{cases} \frac{\alpha + N_i}{k^\circ \alpha + N} C(D, L) & \text{if } i \in \Sigma^\circ \\ \frac{1}{n - k^\circ}(1 - C(D, L)) & \text{if } i \notin \Sigma^\circ \end{cases}$$

*where*

$$C(D, L) = \sum_{k = k^\circ}^{L} \frac{k^\circ \alpha + N}{k\alpha + N} P(k \mid D).$$

*Moreover,*

$$P(S = k \mid D) = \frac{m_k}{\sum_{k' \geq k^\circ} m_{k'}},$$

*where*

$$m_k = P(S = k) \frac{k!}{(k - k^\circ)!} \cdot \frac{\Gamma(k\alpha)}{\Gamma(k\alpha + N)}$$

*and* $\Gamma(x) = \int_0^\infty t^{x-1} e^{-t} dt$ *is the* gamma *function. Thus,*

$$C(D, L) = \frac{\sum_{k=k^\circ}^{L} \frac{k^\circ \alpha + N}{k\alpha + N} m_k}{\sum_{k' \geq k^\circ} m_{k'}}.$$

We can think of $C(D, L)$ as scaling factor that we apply to the Dirichlet prediction that assumes that we have seen all of the feasible symbols. The quantity $1 - C(D, L)$ is the probability mass assigned to *novel* (i.e., unseen) outcomes.

In some of the methods discussed above we need to sample a parameter vector from a sparse-multinomial prior. Probable parameter vectors according to such a prior are sparse, i.e., contain few non-zero entries. The choice of the non-zero entries among the outcomes that were not observed is done with uniform probability. This presents a complication since each sample will depend on some unobserved states. To "smooth" this behaviour we sample from the distribution over $V^\circ$ combined with the novel event. We sample a value of $k$ from $P(S = k|D)$. We then, sample from the Dirichlet distribution of dimension $k$ where the first $k^\circ$ elements are assigned hyper-parameter $\alpha + N_i$, and the rest are assigned hyper-parameter $\alpha$. The sampled vector of probabilities describes the probability of outcomes in $V^\circ$ and additional $k - k^\circ$ events. We combine these latter probabilities to be the probability of the novel event.

## References

Andre, D., Friedman, N. & Parr, R. (1997), Generalized prioritized sweeping, *in* 'Advances in Neural Information Processing Systems', Vol. 10.

Bishop, C. M. (1995), *Neural Networks for Pattern Recognition*, Oxford University Press, Oxford.

Dearden, R., Friedman, N. & Russell, S. (1998), Bayesian Q-learning, *in* 'Proceedings of the Fifteenth National Conference on Artificial Intelligence (AAAI-98)'.

Degroot, M. H. (1986), *Proability and Statistics*, 2nd edn, Addison-Wesley, Reading, Mass.

Friedman, N. & Singer, Y. (1999), Efficient bayesian parameter estimation in large discrete domains, *in* 'Advances in Neural Information Processing Systems 11', MIT Press, Cambridge, Mass.

Heckerman, D. (1998), A tutorial on learning with Bayesian networks, *in* M. I. Jordan, ed., 'Learning in Graphical Models', Kluwer, Dordrecht, Netherlands.

Howard, R. A. (1966), 'Information value theory', *IEEE Transactions on Systems Science and Cybernetics* **SSC-2**, 22–26.

Kaelbling, L. P., Littman, M. L. & Moore, A. W. (1996), 'Reinforcement learning: A survey', *Journal of Artificial Intelligence Research* **4**, 237–285.

Kanazawa, K., Koller, D. & Russell, S. (1995), Stochastic simulation algorithms for dynamic probabilistic networks, *in* 'Proceedings of the Eleventh Conference on Uncertainty in Artificial Intelligence (UAI-95)', Morgan Kaufmann, Montreal.

Kearns, M. & Singh, S. (1998), Near-optimal performance for reinforcement learning in polynomial time, *in* 'Proceedings of the Fifteenth Int. Conf. on Machine Learning', Morgan Kaufmann.

Koller, D. & Fratkina, R. (1998), Using learning for approximation in stochastic processes, *in* 'Proceedings of the Fifteenth International Conference on Machine Learning', Morgan Kaufmann, San Francisco, Calif.

Moore, A. W. & Atkeson, C. G. (1993), 'Prioritized sweeping—reinforcement learning with less data and less time', *Machine Learning* **13**, 103–130.

Ripley, B. D. (1987), *Stochastic Simulation*, Wiley, NY.

Russell, S. J. & Wefald, E. H. (1991), *Do the Right Thing: Studies in Limited Rationality*, MIT Press, Cambridge, Mass.

Sutton, R. S. (1990), Integrated architectures for learning, planning, and reacting based on approximating dynamic programming, *in* 'Proceedings of the Seventh Int. Conf. on Machine Learning', Morgan Kaufmann, pp. 216–224.